\documentclass[reprint,pre,aps,superscriptaddress,amsmath,floatfix,amssymb]{revtex4-1} 
\usepackage[utf8]{inputenc}
\usepackage[T1]{fontenc}
\usepackage{CJK}
\usepackage{graphicx}   
\usepackage{dcolumn}    
\usepackage{bm}         
\usepackage{hyperref}   
\usepackage{xcolor}
\usepackage{subfigure}
\usepackage{ulem}
\usepackage{mathtools}

\definecolor{mygreen}{RGB}{26, 148, 49}

\newcommand*\Laplace{\mathop{}\!\mathbin\bigtriangleup}
\newcommand{\given}[1][]{\,#1\lvert\,}

\graphicspath{{./Images/}} 

\begin{document}

\title{Cascade of Phase Transitions for Multi-Scale Clustering}
\author{Tony Bonnaire}
\affiliation{Université Paris-Saclay, CNRS, Institut d'astrophysique spatiale, 91405, Orsay, France.}
\affiliation{Université Paris-Saclay, TAU team INRIA Saclay, CNRS, Laboratoire de recherche en informatique, 91190, Gif-sur-Yvette, France.}

\author{Aurélien Decelle}
\affiliation{Université Paris-Saclay, TAU team INRIA Saclay, CNRS, Laboratoire de recherche en informatique, 91190, Gif-sur-Yvette, France.}
\affiliation{Departamento de Física Téorica I, Universidad Complutense, 28040 Madrid, Spain}

\author{Nabila Aghanim}
\affiliation{Université Paris-Saclay, CNRS, Institut d'astrophysique spatiale,
91405, Orsay, France.}

\date{\today}

\begin{abstract}
We present a novel framework exploiting the cascade of phase transitions occurring during a simulated annealing of the Expectation-Maximisation algorithm to cluster datasets with multi-scale structures. Using the weighted local covariance, we can extract, \textit{a posteriori} and without any prior knowledge, information on the number of clusters at different scales together with their size.
We also study the linear stability of the iterative scheme to derive the threshold at which the first transition occurs and show how to approximate the next ones.
Finally, we combine simulated annealing together with recent developments of regularised Gaussian mixture models to learn a principal graph from spatially structured datasets that can also exhibit many scales.
\end{abstract}

\maketitle


\section{Introduction}

Many optimisation and inference problems have been shown to have an equivalent formulation in statistical physics \citep{Rose1998, Mezard2009a} that allowed a brand-new look at some long-standing problems and improved the understanding of complex systems \citep{Mezard2009b, Decelle2011}. In particular, the identification of the phase diagram of a model can bring interesting new insights such as knowing if a given information can be retrieved depending on the model's parameters and the dataset at hand. In the context of clustering, it has been demonstrated that the Gaussian Mixture Model (GMM) can be formulated as a statistical mechanics problem \citep{Rose1990, Kappen2000} where the negative log-likelihood can be interpreted as a free energy.

GMMs are extensively used in many fields of science like physics, statistics and machine learning, mainly for complex density estimation \citep{Li2000} or for unsupervised tasks like clustering \citep{Jain2000}. Some attempts to solve other classes of NP-hard problems such as the travelling salesman \citep{Yuille1990} or to model principal curves as an extension of principal components \citep{Tibshirani1992} were also conducted. Standard methods to fit model parameters rely on the Expectation-Maximisation (EM) algorithm \citep{Dempster1977}, an iterative procedure maximising the log-likelihood with guaranteed convergence toward a local maximum \citep{Wu1983}. However, the direct application of EM algorithm is known to be easily trapped in local maxima of multi-modal likelihoods leading to variability in the provided results depending on the initialisation of the algorithm \citep{Kloppenburg1997, Ueda1998}. This problem, coupled with the main drawback of mixture modelling, i.e. the choice of the number of components to model the data, make parametric mixture models very sensitive to the initialisation and the choice of hyper-parameters. In that regard, the statistical physics formulation of the clustering problem helped to overcome these issues by making use of deterministic simulated annealing allowing the relaxation of the non-convex optimisation problem by solving it iteratively while a parameter, assimilated to a temperature, in our case, the variance of all components, is slowly reduced \cite{Kirkpatrick1983}.

Following these steps, we aim at showing how the latter formulation can be useful to understand and analyse the outcome of GMMs. In particular, we exploit the cascade of phase transitions occurring during annealing procedures of the EM algorithm to build a hierarchical multi-scale description of a dataset. By defining an overlap between the ground truth and the inferred partitions, we show on artificial datasets how it can be interpreted as an order parameter whose value follows the sequence of phase transitions. In more general cases, where the ground truth is not known, we use a physical observable to get an \textit{a posteriori} interpretation of what happened during the annealing.
For a given mixture component, we track the hierarchy of scales that are being represented through the eigenvalues of the weighted covariance matrix and use it to extract information on the structure, i.e. the number of clusters, their scale and their hierarchy without prior information on the ''correct'' number of clusters, even in high dimension when direct visualisation is impossible.
To represent different scales simultaneously, we propose an alternative to the classical annealing in which the temperature is the mode of a prior distribution on variances.
We show that the threshold at which the first transition occurs can be computed exactly from the linear stability of the fixed-point iterative scheme in both cases. Approximate positions of successive thresholds can also be estimated when the dataset is clustered into several sub-systems. The proposed methodology hence leads to a representation: i) independent from the inputted number of components, ii) independent from the data space dimensions, iii) hierarchically nested from high scales to lower ones.

In Section \ref{sect:GMM}, we introduce the mathematical background of GMMs and review the main equations of EM algorithm to estimate parameters of the model. Section \ref{sect:Hard} presents the hard annealing setting and motivates the use of the procedure to learn hierarchical representations. Section \ref{sect:Soft} focuses on the interest of combining a regularised version of the GMM with the annealing procedure for the study of complex datasets and provides theoretical derivations of the threshold at which the first transition occurs. Finally, in Section \ref{sect:Graph}, we illustrate how these transitions can be used together with the recently introduced graph regularisation of Gaussian mixtures (GMs) to learn a principal graph structure from spatially structured datasets, even with high heteroscedasticity of the sampling.


\section{Gaussian Mixture Models and Expectation-Maximisation algorithm} \label{sect:GMM}

It is not always possible to fit known probability distributions to those resulting from a set of physical measurements $\boldsymbol{X} = \{\boldsymbol{x}_i\}_{i=1}^N$ with $\boldsymbol{x}_i \in \mathbb{R}^{D}$. GMMs naturally allow the modelling of complex probability distributions as a linear combination of $K$ Gaussian components with unknown parameters. In what follows, we restrict ourselves to spherical and uniform GMs with means $\{\boldsymbol{\mu}_k\}_{k=1}^K$ and variances $\{\sigma_k^2\}_{k=1}^K$ such that
\begin{equation} \label{eq:model}
    p(\boldsymbol{x}_i \given \boldsymbol{\Theta}) = \frac{1}{K} \sum_{k=1}^{K} \mathcal{N}(\boldsymbol{x}_i, \boldsymbol{\theta}_k),
\end{equation}
with $\boldsymbol{\Theta} = \{\boldsymbol{\theta}_0, \ldots, \boldsymbol{\theta}_k\}$ the set of parameters with $\boldsymbol{\theta}_k = \left(\boldsymbol{\mu}_k, \sigma_k\right)$.

EM states that the log-likelihood can be maximised through an iterative procedure involving two alternating steps by considering a set of latent variables $\{z_i\}_{i=1}^N$ with $z_i \in \{0, \ldots, K\}$ describing the label of the cluster that generated the datapoint $\boldsymbol{x}_i$. The E-step first estimates the posterior distribution of the latent variables from the current values of parameters as
\begin{equation} \label{Estep}
    p_{ik} = p(z_i \given \boldsymbol{x}_i, \boldsymbol{\theta}_k) = \frac{\displaystyle \exp \left( - \lVert \boldsymbol{x}_i - \boldsymbol{\mu}_k \rVert^2_2 / 2\sigma_k^2 \right)}{ \displaystyle \sum_{j=1}^K\exp \left( - \lVert \boldsymbol{x}_i - \boldsymbol{\mu}_j \rVert^2_2 / 2\sigma_j^2 \right)}.
\end{equation}
The M-step then refines parameter estimates based on the current values of $p_{ik}$ as
\begin{equation} \label{Mstep}
    \left\{
    \renewcommand*{\arraystretch}{3.0}
    \begin{array}{l}
        \displaystyle \boldsymbol{\mu}_k = \frac{\sum_{i=1}^N p_{ik} \boldsymbol{x}_i}{N_k}, \\
        \sigma_k^2 = \displaystyle \frac{\sum_{i=1}^N p_{ik} \lVert \bm{x}_i - \bm{\mu}_k \rVert_2^2}{DN_k},
    \end{array}
    \right.
\end{equation}
where $N_k = \sum_{i=1}^N p_{ik}$.

From the statistical physics point of view, identical update equations can be obtained by deriving the free energy under a quadratic cost function for the datapoints assignation to clusters \citep{Rose1998}.


\section{Phase transitions in hard annealing} \label{sect:Hard}

Simulated annealing in EM procedure was introduced to overcome the issue of local maxima. It consists in reducing iteratively the variance of all components such that $\forall k \in \{1, \ldots, K\}, \sigma_k^2 = \sigma^2$, where $\sigma^2$ is the controlled value of the variance during the annealing. The idea behind these approaches is to smooth the likelihood by starting with a very high variance leading to a concave function. Decreasing it slowly leads to a finer and finer description of the dataset hence resulting in a more complex likelihood function with multiple modes appearing.

In this work, we focus on a particular aspect of the annealing based on successive phase transitions.
First, we are interested in the range of $\sigma^2$ values for which the likelihood is concave and hence all of the $K$ components are collapsed into a single location centred at $1/N\sum_{i=1}^N \boldsymbol{x}_i$.
This critical quantity, noted $T_c^\text{hard}$, is known to be the maximum eigenvalue of the data covariance matrix \citep{Rose1990}. When $\sigma^2 > T_c^\text{hard}$, even though $K$ components are used in the model, they are all collapsed as $K_r = 1$ physical cluster at the centre of mass (c.m.) of the dataset. When $\sigma^2$ becomes slightly smaller than $T_c^\text{hard}$, the likelihood is deformed and centres get aligned with the first principal direction given by the data covariance matrix $\boldsymbol{C}$. When $\sigma^2$ continues to decrease, the dataset description becomes more and more detailed and $K_r$ takes increasing values.

We propose a novel way to extract information on the structure of a dataset from the annealing process by tracking the evolution of the size represented by a given component $k$ through the maximum eigenvalue $\Gamma_k$ of its weighted covariance matrix, namely $\boldsymbol{\Sigma}_k = 1/N_k \sum_{i=1}^N p_{ik} (\boldsymbol{x}_i - \boldsymbol{\mu}_k)^\text{T}(\boldsymbol{x}_i - \boldsymbol{\mu}_k)$.

\begin{figure}
    \centering

    \subfigure{\includegraphics[width=0.90\linewidth]{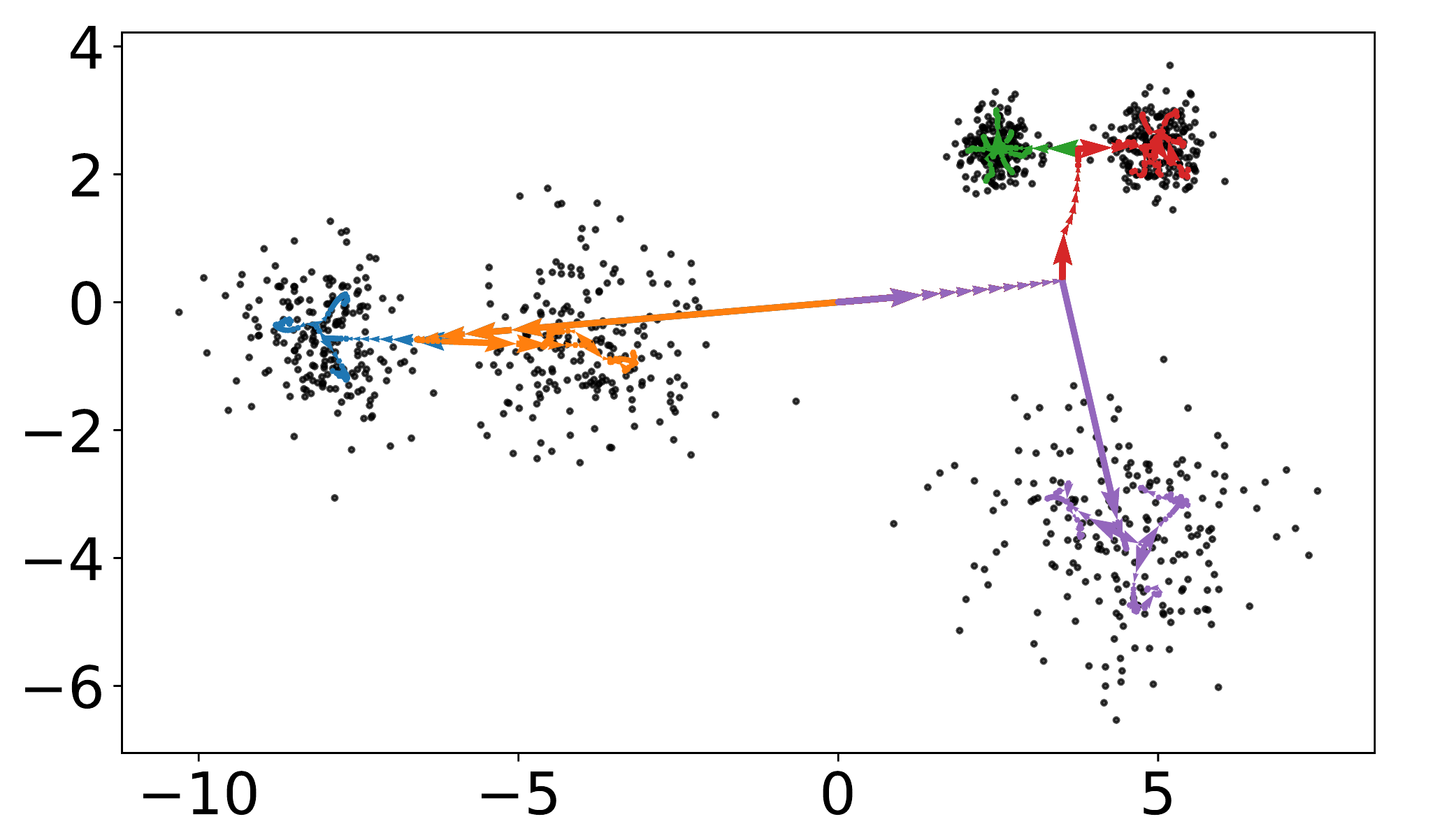}}
    
    \vspace{-0.35cm}
    
    \subfigure{\includegraphics[width=1\linewidth]{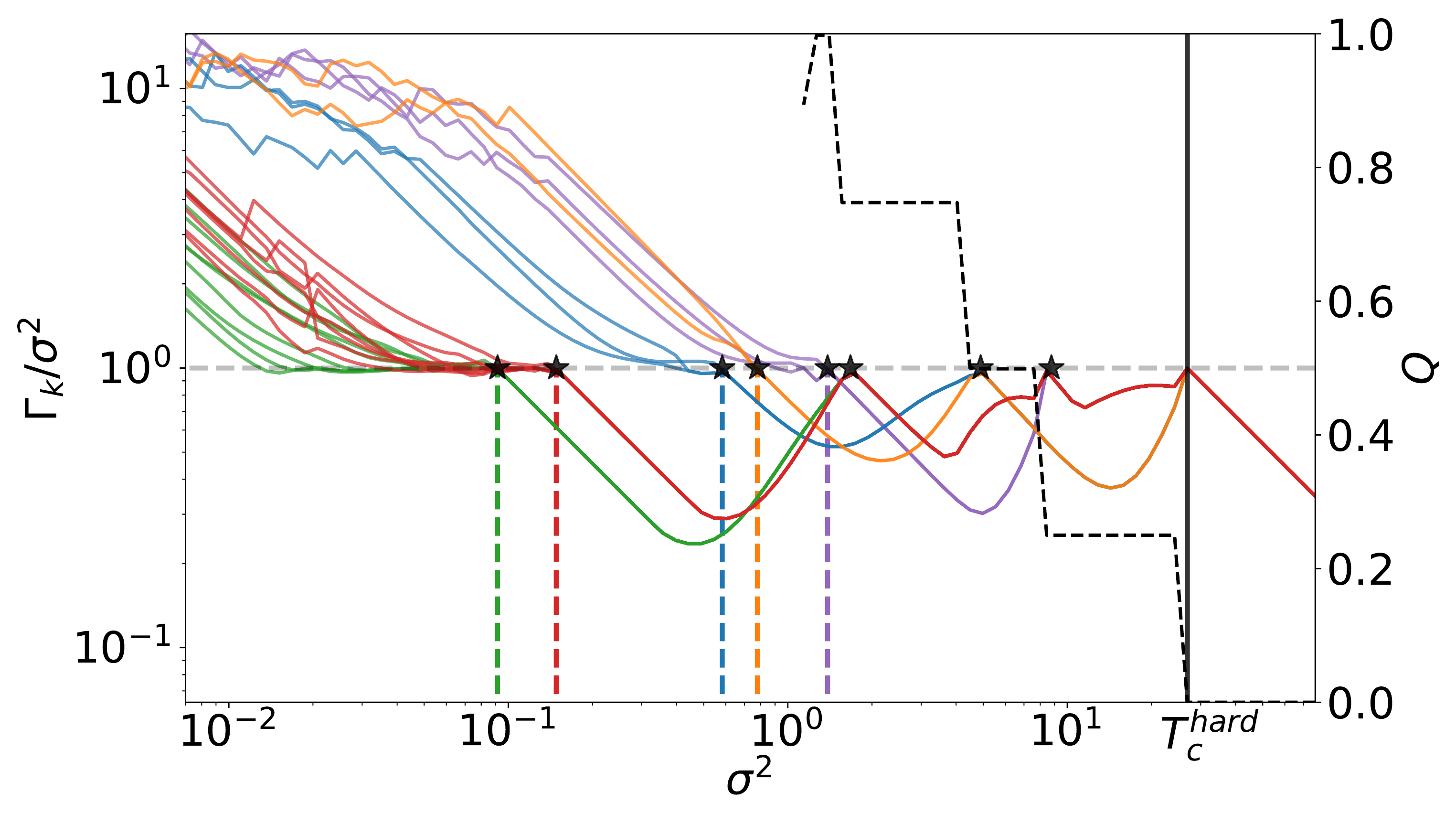}}
    
    \caption{\textit{(top)} Displacement of $K=25$ centres during the annealing procedure for a dataset with five spherical Gaussian clusters. Colours indicate in which final cluster the centre ends. \textit{(bottom)} Evolution of the ratio $\Gamma_k/\sigma^2$ as a function of $\sigma^2$. Black stars correspond to the scales of successive transitions, the black vertical line to $T_c^\text{hard}$ and coloured ones indicate the size of the clusters as defined by the maximum eigenvalue of the empirical covariance. The black dashed curve shows to the evolution of $Q$ as defined in Eq. \eqref{order_parameter} that we identify as an order parameter. This quantity is not represented for $\sigma \leq 1$ since the number of physical clusters $K_r$ begins to be higher than the number of generated clusters $q$.}
    
    \label{fig:Rose_MS_Hard}
\end{figure}

Figure \ref{fig:Rose_MS_Hard} illustrates the evolution of the ratio $\Gamma_k/\sigma^2$ during the annealing for an artificial dataset with five clusters and $K=25$ centres coloured by their end-point cluster. In the bottom panel, we see the cascade of transitions and successive splitting of centres when $\sigma^2$ decreases. When two or more centres collapse, they share similar values of $\boldsymbol{\mu}_k$ and $p_{ik}$ leading to similar evaluations of $\Gamma_k$. This is why all lines are superimposed for $\sigma^2 > T_c^\text{hard}$. Each time a curve reaches the horizontal unit line, one of the $K_r$ sub-system made of collapsed components reached the temperature of the sub-dataset it represents, namely $\sigma^2 \simeq \Gamma_k$. From there, we observe either a bounce or a cross of the line. When bouncing, there is a split between two populations of centres that were representing the same part of the dataset but that will take different paths. Centres thus move toward a smaller cluster and the value of $\Gamma_k$ decreases. Crossing the line occurs when centres split inside an individual cluster due to its inner random structure. In that case, the imposed variance gets smaller than the physical one. Since the transitions are driven by the maximum eigenvalues of the empirical covariance matrices, it is this quantity that is plotted as vertical lines for the size of spherical clusters on the figures. Note that if we use instead the empirical variance, this estimates would be slightly shifted to lower values and the transition would occur earlier in the annealing than the true variance of the cluster.

Following \textit{a posteriori} the several curves and the successive transitions provide an informative insight on the structure of the dataset. It also allows to visualise the evolution of the local size representation of the data and the interactions between centres. In the bottom panel of Fig. \ref{fig:Rose_MS_Hard}, we clearly see that $\{\text{purple, red, green}\}$ sets of centres represent the same information when $\sigma^2 > 9$ and then split into $\{\text{purple}\}$ and $\{\text{red, green}\}$. This indicates the presence of a sub-system of two clusters. Later, we observe a crossing of the horizontal line for the $\{\text{purple}\}$ centres before splitting again after crossing. This indicates that the effective variance of the cluster is larger than the one fixed by the annealing and, therefore, that these centres now describe fluctuations within a ''true'' cluster.  The $\{\text{red, green}\}$ sets of centres split at lower $\sigma^2$ followed as well by a crossing of the line at the scale of individual cluster sizes.

Successive transitions can be computed in two steps: First by identifying the $K_r$ macro-components resulting from the collapse of centres based on their positions and then by assigning to each data point the label of the macro-component that most probably generated it. We are thus assuming, at a given iteration, a GMM with $K_r$ components to compute responsibilities from Eq. \eqref{Estep}. Hence, we can group data points with identical labels and compute the next transition as the maximum eigenvalue of the covariance matrix for each sub-system. These quantities, basically corresponding to successive evaluations of the critical temperature in sub-systems, can be approximated during the annealing and are shown as black stars on the bottom panel of Fig. \ref{fig:Rose_MS_Hard}. 

We identify the overlap $Q$, defined as the quality of the data classification at each temperature, as an order parameter whose value changes throughout the several phases during the annealing. Formally,
\begin{equation}    \label{order_parameter}
    Q\left( \{\hat{z}_i\}, \{z_i\}\right)= \frac{\operatorname*{max}_{\pi \in \Pi} \frac{1}{N} \sum_i \delta_{\hat{z}_i, \pi \left(z_i\right) }- 1/q}{1 - 1/q},
\end{equation}
where $\delta$ is the Kronecker delta, $\Pi$ denotes all the possible permutations of the set $\{1, \ldots, q\}$ with $q$ the true number of clusters used to generate the data and $\hat{z}_i = \operatorname*{argmax}_k p_{ik}$ the estimated latent variable for the affiliation of the datapoint $\boldsymbol{x}_i$. $Q$ is hence defined such that when $\forall i, \hat{z}_i = z_i$, $Q=1$ and for a random assignation $p_{ik} = 1/K$, $Q=0$. By doing so, $Q$ is zero when $\sigma^2 > T_c^\text{hard}$ and undergoes successive transitions as the system is cooled down, as illustrated on the bottom panel of Fig. \ref{fig:Rose_MS_Hard}. During the annealing, $Q$ remains at $1$ for the range of $\sigma^2$ between the last split of centres between two true clusters and before the first split due to the inner random structure of one of them. Note that this metric is not applicable when $K_r > q$ since the dataset is partitioned into more clusters than actually used for the generation, and this is why the curve is not shown for $\sigma^2 \leq 1$.


\begin{figure}
    \centering
    \includegraphics[width=1\linewidth]{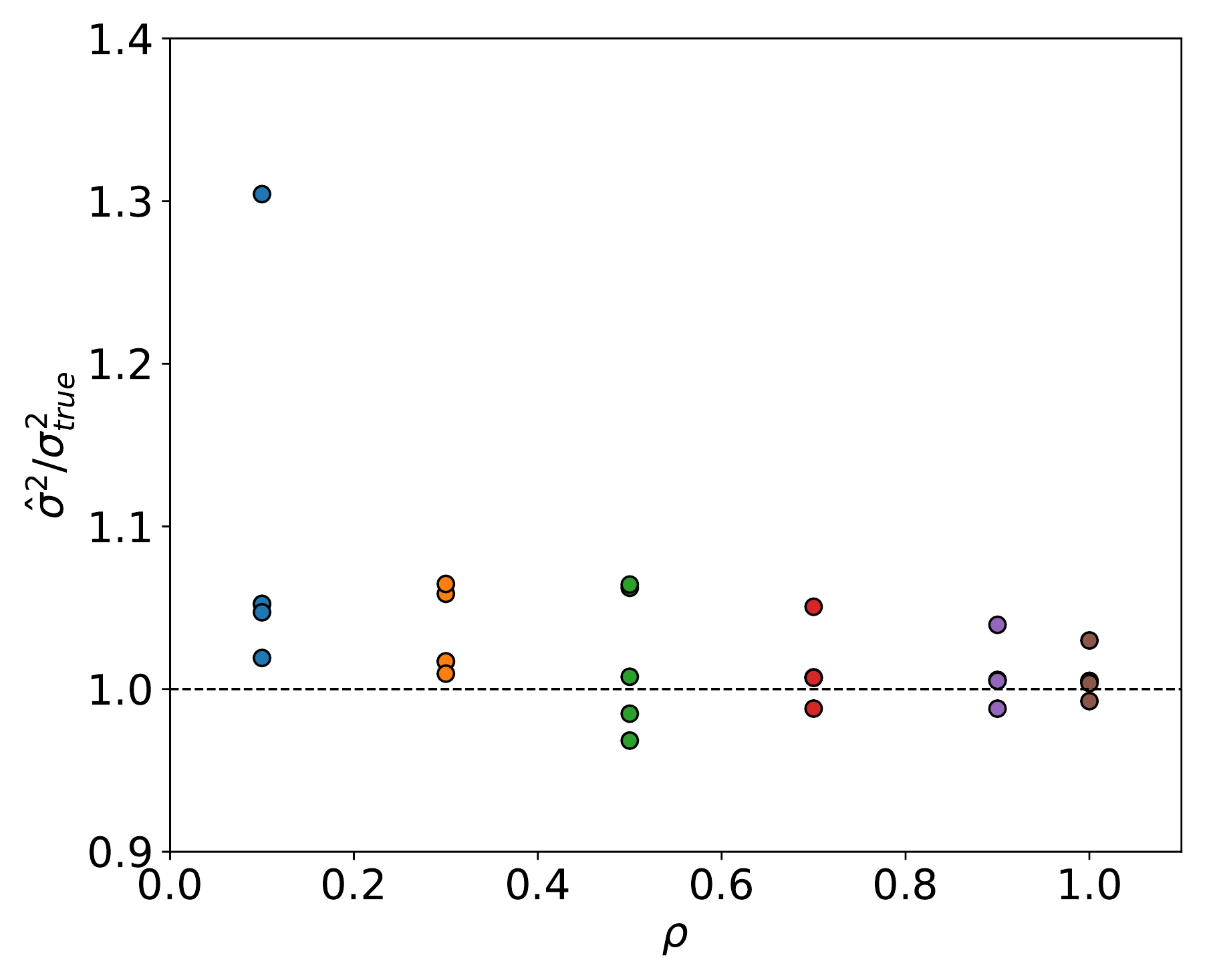}
    
    \caption{Ratio between the estimated variances $\hat{\sigma}^2$ obtained when freezing the $K=25$ centres when $\Gamma_k / \sigma^2 \simeq 1$ and the empirical ones $\sigma^2_\text{true}$ from data of Fig. \ref{fig:Rose_MS_Hard}. Colours refers to several value of $\rho$, the proportion of datapoints left for the computation.}
    
    \label{fig:Robustness}
\end{figure}

To further assess the robustness and accuracy of the transitions, we use the dataset from Fig. \ref{fig:Rose_MS_Hard} where only a fraction $\rho$ of the datapoints is randomly kept for the computation. During the annealing, we freeze all the $K=25$ centres at the last split before reaching $\Gamma_k / \sigma^2 = 1$ and then let variances evolve freely hence providing an estimate for each detected cluster that we note $\hat{\sigma}^2$. Figure \ref{fig:Robustness} shows that the retrieved variances are, even in highly sparse sampling settings, with $\rho \leq 30\%$, close to the true ones of the clusters. It is worth emphasising that all five clusters are always correctly identified and that the value of $Q$ is always close to $1$ at the end of the process, showing the ability of the method to highlight structures, even in sparse configurations.

\bigskip

\begin{figure}
    \centering
    \subfigure{\includegraphics[width=0.88\linewidth]{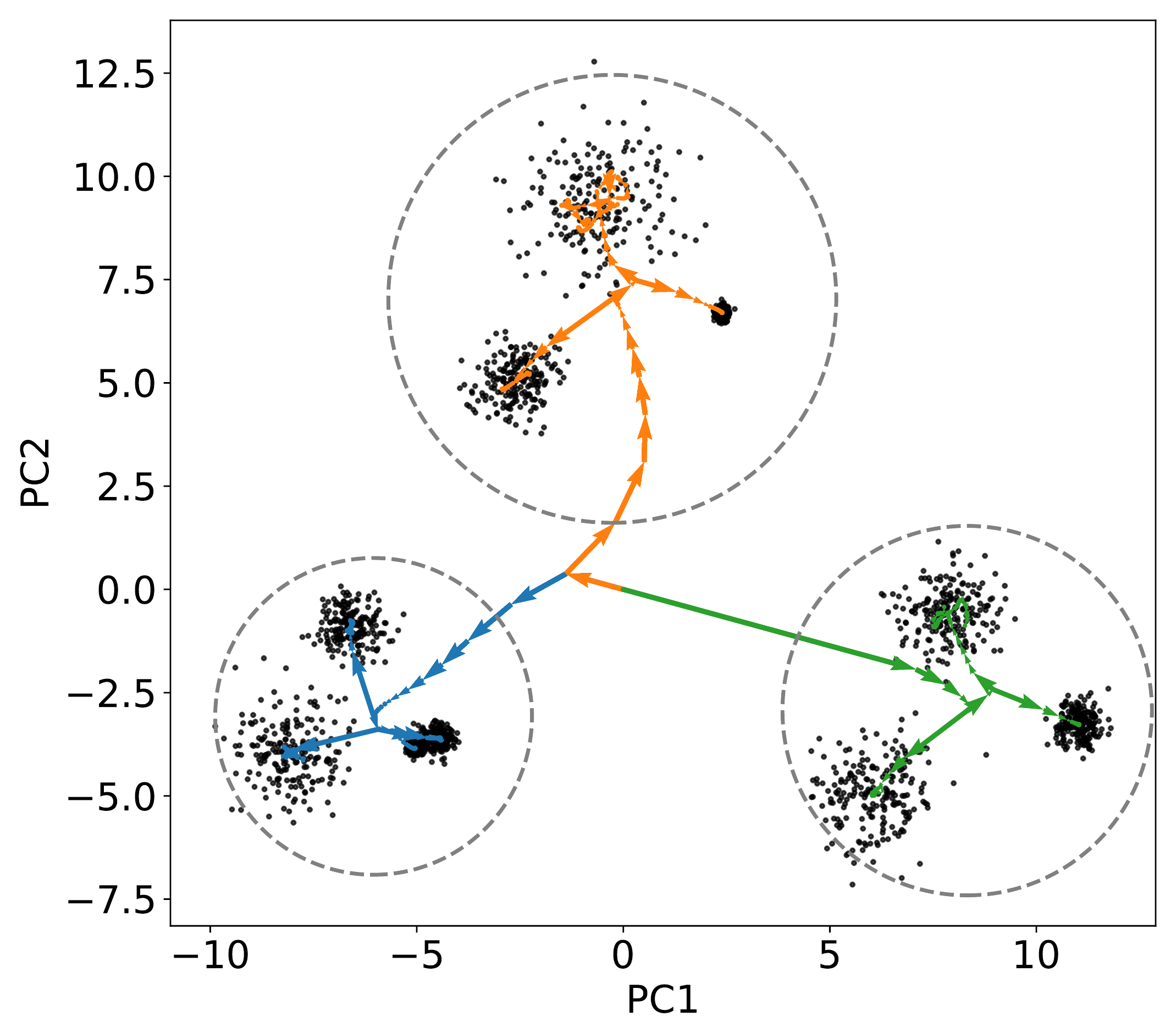}}
    \vspace{-0.35cm}
    \subfigure{\includegraphics[width=0.88\linewidth]{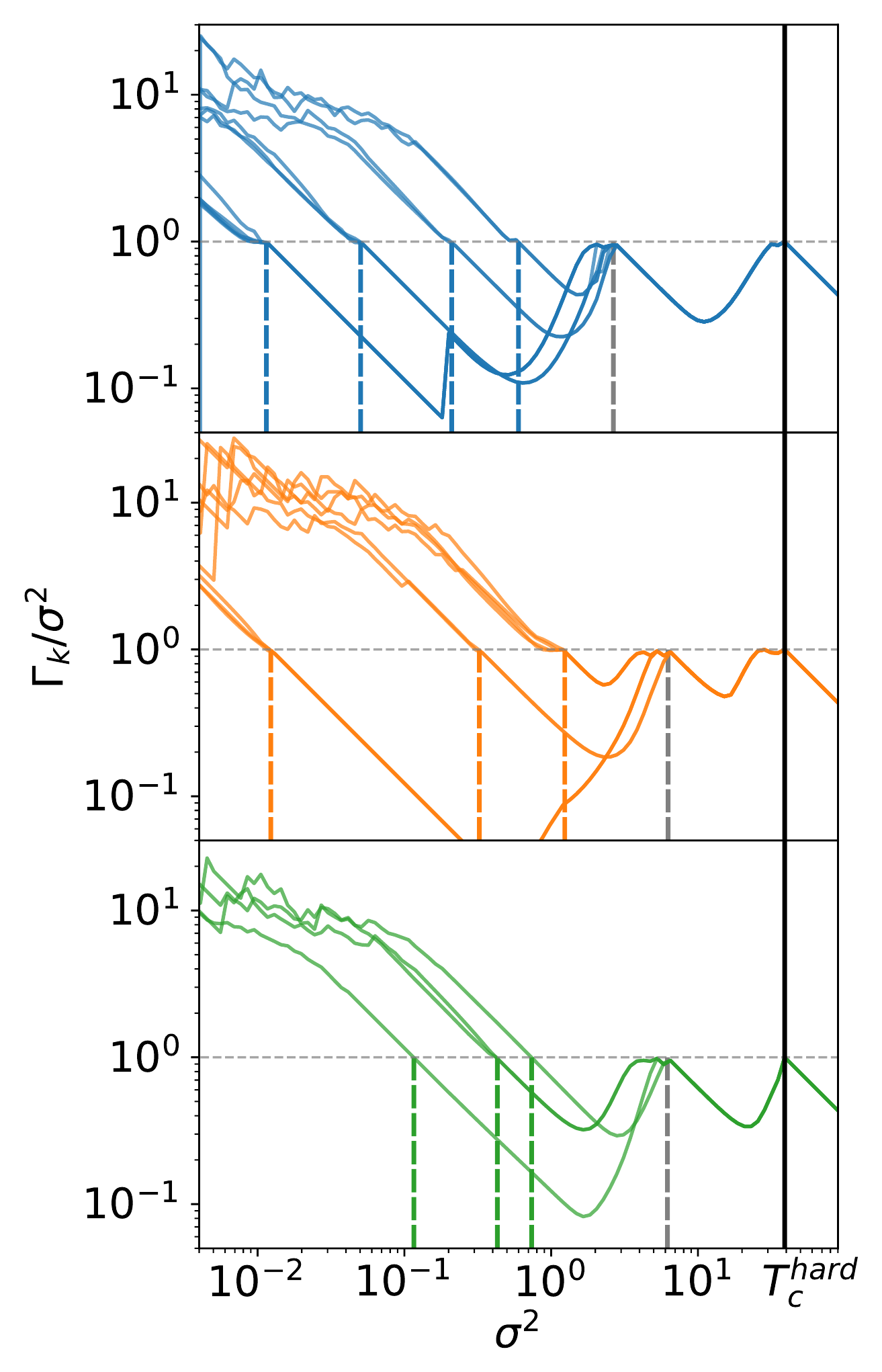}}
    
    \caption{\textit{(top)} Displacement of $K=25$ centres during the annealing procedure for a dataset made of ten 5D spherical Gaussian clusters visualised in the plane of the two first principal components. Colours depend on the macro-cluster the component stands in at the last iteration. \textit{(bottom)} Evolution of the ratio $\Gamma_k/\sigma^2$ as a function of $\sigma^2$, the hard annealing parameter. Coloured vertical lines indicate the actual size of the corresponding Gaussian cluster or macro-cluster (grey lines) as defined by the maximum eigenvalue of the empirical covariance.
    }
    \label{fig:MS_Hard}
\end{figure}

Usual applications of GMMs for clustering are performed blindly by inputting the desired $K$ to obtain a classification making use of all components. Some criteria, based on information theory \citep{Akaike1974, Oliver1996} or Bayesian approaches \citep{Schwarz1978, Roeder1997}, were proposed to overcome this major drawback of unsupervised clustering. Here, we propose an approach to avoid such a $K$-dependent unique solution. A key aspect of the annealing is the collapse of $\{\boldsymbol{\mu}_k\}_{k=1}^K$ at the c.m. of successive sub-datasets providing a hierarchical view of clustering with an increasing number of physical clusters. The proposed diagram enables to capture this set of nested representations as shown in Fig. \ref{fig:MS_Hard} for a 5D artificial dataset made of ten Gaussian clusters, spatially appearing as three clusters at larger scale. Transitions occurring at large scales in each panel (grey vertical lines) clearly indicate that the dataset is described as three different physical clusters. Pursuing the decrease of variances leads to a finer description where each of the three macro-clusters splits into smaller ones that still have physical interpretations \footnote{The separation in three panels with different colours is for visualisation purposes. No prior knowledge was used in the analysis.}. Such information on the spatial organisation of the dataset are of crucial importance when having no prior idea of its structure nor the number of underlying components.

\section{Phase transitions in soft annealing} \label{sect:Soft}

When the dataset is more complex with nested structures or overlapping clusters of different sizes, the previously presented analysis is not suitable. Multiple scales cannot be represented at the same time, hence biasing those of embedded structures toward higher values.
To overcome this, we rely on a modified annealing acting on the mode of an \textit{a priori} distribution on each $\sigma_k^2$. More particularly, we use the conjugate prior for variances, namely an inverse-Gamma distribution, with shape parameter $1+\lambda_{\sigma}$ and scale parameter $\lambda_{\sigma} \sigma^2$ so that the distribution has a mode at $\sigma^2$. Formally, it reads
\begin{equation}
\log p(\sigma^2_k) = - \lambda_{\sigma} \left[ \log \sigma_k^2 + \frac{\sigma^2}{\sigma_k^2} \right] + \text{cte},
\end{equation}
where the constant comes from the normalisation of the probability distribution.
Introducing such a prior modifies the update Eq. \eqref{Mstep} for variances as
\begin{equation} \label{eq:Soft_update}
    \sigma_k^2 = \displaystyle \frac{\sum_{i=1}^N p_{ik} \lVert \bm{x}_i - \bm{\mu}_k \rVert_2^2 + 4 \lambda_{\sigma} \sigma^2}{D\sum_{i=1}^N p_{ik} + 4\lambda_{\sigma}}.
\end{equation}
Consequently, when $\lambda_{\sigma} \to 0$, the prior, and hence the annealing, has no effect and components update their variances as Eq. \eqref{Mstep}. Inversely, for a large enough value, $\sigma_k^2$ will be close to $\sigma^2$ resulting in the classical annealing procedure. Choosing intermediate values for $\lambda_{\sigma}$ hence imposes a broad trend for all components but lets each of them correct the prior by the actual value of the neighbouring covariance. In what follows, we refer to this procedure as "soft annealing" that we distinguish from the "hard annealing" to describe the classical procedure acting directly on the variance parameter. There is no rule to fix the hyper-parameter $\lambda_{\sigma}$ and, in this work, we adopt $\lambda_{\sigma} = 2$ \footnote{In our experiments, keeping $\lambda_\sigma \approx O(1)$ did not change the results quantitatively.}.

Similarly as in the hard annealing case, we can compute the threshold value $T_c^\text{soft}$ such that all components are collapsed at the c.m. when $\sigma^2 > T_c^\text{soft}$.
Without any loss of generality, the dataset can be considered centred, with $\sum_{i=1}^N \boldsymbol{x}_i = \boldsymbol{0}_D$ where $\boldsymbol{0}_D$ is the $D$-dimensional zero vector. 
In a first step, we propose to derive the fixed-point variance $\sigma_0^2$ of all centres when $\sigma^2 \gg T_c^\text{soft}$. Linearly Taylor expanding the expression of $p_{ik}$ given by Eq. $\eqref{Estep}$ for small perturbations $\boldsymbol{\mu}_k \approx \boldsymbol{0}_D$ leads to
\begin{equation} \label{eq:responsibility_taylor}
    p_{ik} = \frac{1}{K} \left[1 + \frac{1}{\sigma_k^2} \boldsymbol{x}_i^\text{T} \boldsymbol{\mu}_k \right] + o(\boldsymbol{\mu}_k).
\end{equation}
Further, assuming that $\sigma^2 > T_c^\text{soft} \implies \forall k \in \{1, \ldots, K\}, \sigma_k^2 = \sigma_0^2$ and injecting $p_{ik}$ expansion in update Eq. \eqref{eq:Soft_update} gives
\begin{equation}
    \sigma_0^2 = \frac{4\lambda_{\sigma}K \sigma^2 + \sum_{i=1}^N \boldsymbol{x}_i^\text{T} \boldsymbol{x}_i}{ND + 4\lambda_{\sigma} K}.
\end{equation}
This equation links $\sigma_0^2$, the actual variance attributed to all Gaussian components, to $\sigma^2$, the soft annealing parameter and is valid in the large $\sigma^2$ limit. By considering small perturbations $\boldsymbol{\epsilon}_k$ and $\delta_k$, respectively around the fixed points $\boldsymbol{\mu}_k = \boldsymbol{0}_D$ and $\sigma_k^2 = \sigma_0^2$, we can derive the set of equations for the vectorised perturbations $\underline{\underline{\boldsymbol{\epsilon}}} = \left(\bm{\epsilon_0}^\text{T}, \ldots, \bm{\epsilon_K}^\text{T}\right)^\text{T} \in \mathbb{R}^{KD\times 1}$ and $\boldsymbol{\delta} = \left(\delta_0, \ldots, \delta_K \right) \in \mathbb{R}^{K\times 1}$,
\begin{align}
    \left.
    \renewcommand*{\arraystretch}{2.5}
    \begin{array}{l}
    \underline{\underline{\bm{\epsilon}}}^{(t+1)} = \displaystyle \frac{1}{\sigma_0^2}\left( \boldsymbol{U} \otimes \bm{C} \right) \underline{\underline{\bm{\epsilon}}}^{(t)} + \left( \boldsymbol{U} \otimes \bm{a} \right)^\text{T} \bm{\delta}^{(t)}, \\
    \bm{\delta}^{(t+1)} = \left( \boldsymbol{U} \otimes \bm{b}\right) \underline{\underline{\bm{\epsilon}}}^{(t)} + c \, \boldsymbol{U} \bm{\delta}^{(t)},
    \end{array}
    \right.
\end{align}
where $\otimes$ denotes the Kronecker product, $\boldsymbol{U} = \left(\boldsymbol{I}_K - \frac{1}{K} \boldsymbol{J}_K \right)$ with $\boldsymbol{I}_K$ is the $K\times K$ identity matrix, $\boldsymbol{J}_k$ the $K\times K$ all-ones matrix, and
\begin{align}
    \bm{a} &= \sum_i \frac{\lVert \boldsymbol{x}_i \rVert^2_2 \bm{x}_i^T}{2N \sigma_0^4}, \\
    \bm{b} &= \sum_i \frac{\lVert \boldsymbol{x}_i \rVert^2_2 \bm{x}_i^T}{m \sigma_0^2}, \\
    c &= \frac{1}{2 \sigma_0^2 m} \left( \sum_i \frac{\lVert \boldsymbol{x}_i \rVert^4_2 }{\sigma_0^2} - D \sum_i \lVert \boldsymbol{x}_i \rVert^2_2 \right),
\end{align}
with $m = ND + 4\lambda_{\sigma}K$.
Putting it all together leads to the matrix representation of the system's perturbations $\underline{\underline{\bm{\eta}}} = \left( \underline{\underline{\bm{\epsilon}}}, \boldsymbol{\delta} \right) \in \mathbb{R}^{K(D+1)\times 1}$
\begin{equation}
    \underline{\underline{\bm{\eta}}}^{(t+1)} = \left( \boldsymbol{U} \otimes \bm{M} \right) \underline{\underline{\bm{\eta}}}^{(t)},
\end{equation}
with $\boldsymbol{M}$ the squared block matrix of order $D+1$
\begin{equation}
\arraycolsep=2.2pt\def\arraystretch{1.8}
\boldsymbol{M} = \sbox0{$\begin{matrix} \boldsymbol{C}/\sigma_0^2 & \boldsymbol{C}/\sigma_0^2\end{matrix}$}
    \left(\begin{array}{c|c}
    \vphantom{\begin{matrix} \boldsymbol{C}/\sigma_0^2 \\ \boldsymbol{C}/\sigma_0^2\end{matrix}} \boldsymbol{C}/\sigma_0^2 & \bm{a}^\text{T} \\
    \hline
    \mathmakebox[\wd0]{\bm{b}} & c
    \end{array}\right),
\end{equation}
where $\boldsymbol{C}$ is the data covariance matrix.

Since the eigenvalues of the Kronecker product are given by the product of all individual eigenvalues of the two matrices involved and that $\boldsymbol{U}$ has eigenvalues $0$ or $1$, we can only restrict the analysis to those of $\bm{M}$. Therefore, the value of $\sigma^2$ at which the first transition occurs, namely $T_c^\text{soft}$, can be derived as the value of $\sigma^2$ such that the spectral radius of $\boldsymbol{M}$ is 1, leading to instabilities in the dynamic of the system.

    
    

\begin{figure}
    \centering
    \subfigure{\includegraphics[width=0.95\linewidth]{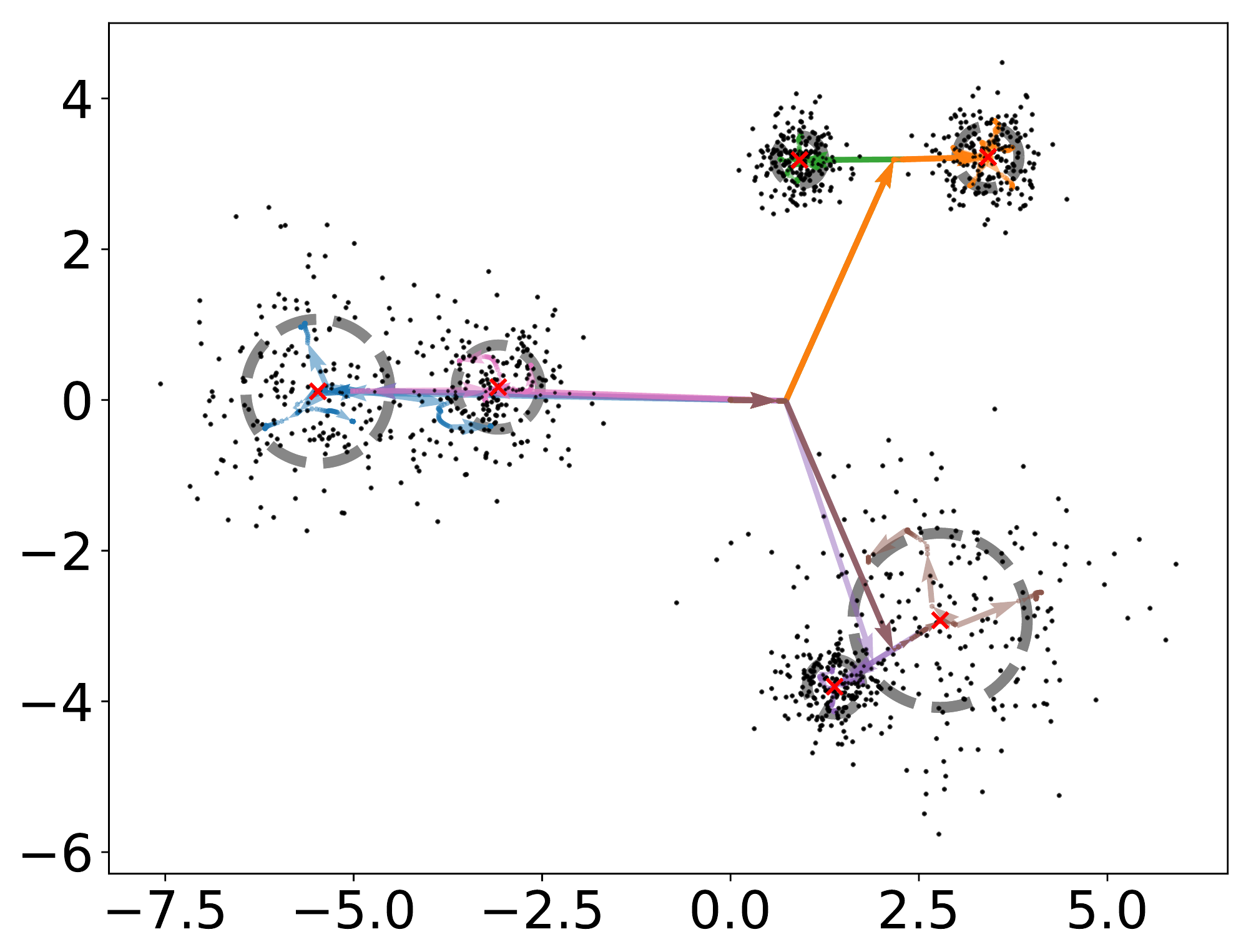}}
    
    
    \subfigure{\includegraphics[width=1\linewidth]{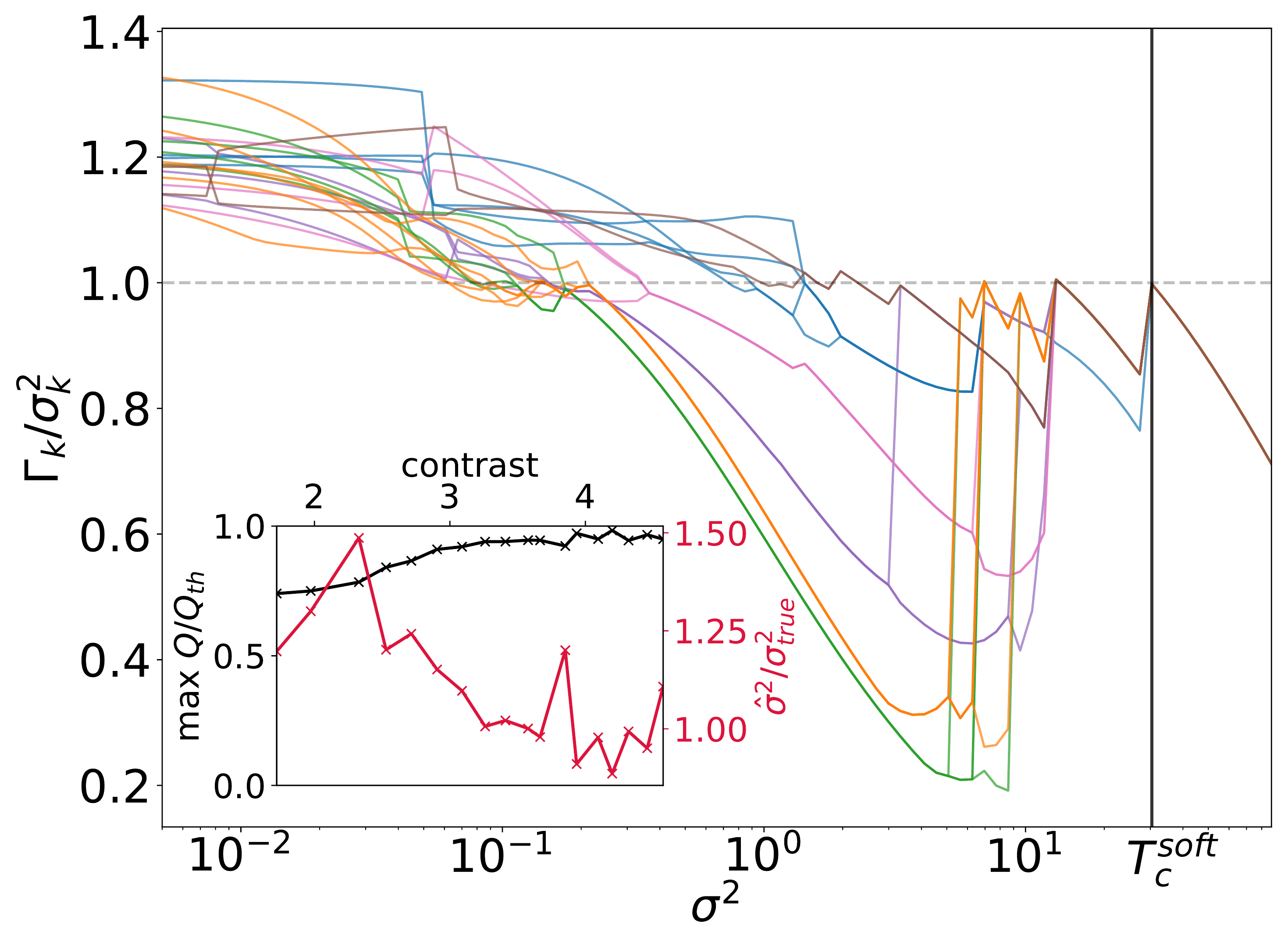}}
    
    \caption{\textit{(top)} Arrows indicate the displacement of $K=25$ centres during the soft annealing procedure for a dataset made of six spherical Gaussian clusters (black points). Colours relate to the cluster in which the component ends. Red crosses and grey dashed circles respectively indicate the positions and variances fixed \textit{a posteriori} when the centre undergoes its last split before remaining above the $\Gamma_k/\sigma_k^2 = 1$ line. \textit{(bottom)} Evolution of the ratio $\Gamma_k/\sigma_k^2$ as a function of $\sigma^2$. The vertical black line corresponds to $T_c^\text{soft}$. The inset figure shows the evolution of the ratios $\max Q / Q_{th}$ and $\hat{\sigma}^2/\sigma^2_{\text{true}}$, when varying the contrast between the two nested clusters.}
    
    \label{fig:Rose_MS_Soft}
\end{figure}

Figure \ref{fig:Rose_MS_Soft} illustrates the result of the soft annealing procedure on an artificial dataset made of six clusters similar as Fig. \ref{fig:Rose_MS_Hard} but with more complexity such as overlapping and nested clusters. The bottom panel focuses on the evolution of the ratio between the size of the represented sub-system by a given component and its actual variance, namely $\Gamma_k/\sigma_k^2$. This is the same physical quantity as in the hard annealing case, except that we relax the constraint on $\sigma_k^2$ which is now varying for each component. In this soft configuration, all the $\boldsymbol{\mu}_k$ are collapsed for $\sigma^2 > T_c^\text{soft}$ followed by steep transitions when $\sigma^2$ decreases. 
This relaxed annealing is especially useful for the representation of the two nested clusters. Even though we would learn that those structures are encapsulated, reaching accurate size description for the smallest component would not be possible in hard annealing because its variance would be boosted by neighbouring data points. The top panel illustrates positions and variances of all $K=25$ components when fixing parameters \textit{a posteriori} at the value they had during the annealing at their last transition point just before crossing the line $\Gamma_k/\sigma_k^2 = 1$. Although $K=25$ components are used, we correctly identify $K_r = 7$ physical clusters with their variances and means as indicated by red crosses and grey circles on Fig. \ref{fig:Rose_MS_Hard}.

To assess the robustness of the soft annealing procedure in clustering complex datasets, we focus on a setup restricted to the two nested clusters of Fig. \ref{fig:Rose_MS_Soft} only and dilute the small one by varying its number of sampling points $N$. This translates into a decreasing contrast between the signal to noise ratios $\sigma / \sqrt{N}$ of the two clusters. The inset of Fig. \ref{fig:Rose_MS_Soft} shows the evolution of the ratio between the maximum overlap value $Q$ obtained during the annealing and $Q_{th}$ the theoretical overlap computed using the ground truth parameters as a function of the contrast. It can be seen that both clusters are recovered when the contrast is sufficiently high (above 1.5 in practice) while below, there is not the necessary information for the model to retrieve it. The ratio between the estimated variances is also shown to and we observe an overestimated variance at lower and lower contrast which explains the decreasing $\max Q / Q_{th}$ ratio. This effect can partly be explained by the uniform weights hypothesis being less and less true when the contrast decreases.

\section{Multi-scale principal graphs} \label{sect:Graph}

Recently proposed methods explore the graph regularisation of GMs to learn a smooth graph representation from point-cloud distributions \citep{Mao2017, Bonnaire2020}.
Based on a Gaussian prior acting on component averages, these approaches rely on a measurement of the graph smoothness through the graph Laplacian as 
\begin{equation}
    \lVert \Laplace{\boldsymbol{\mu}} \rVert^2 = \sum_{i=1}^K \sum_{j=1}^K \boldsymbol{A}_{ij} \lVert \boldsymbol{\mu}_i - \boldsymbol{\mu}_k \rVert^2_2,
\end{equation}
where $\boldsymbol{A}$ is the adjacency matrix taking value $1$ when centres $i$ and $j$ are linked and $0$ otherwise.
Exploiting this prior knowledge in the GMM leads to the regularised model 
\begin{equation}
    \log p(\boldsymbol{\Theta} \given \boldsymbol{x}_i) \propto \log p(\boldsymbol{x}_i \given \boldsymbol{\Theta}) + \log p(\boldsymbol{\Theta}),
\end{equation}
with $\log p(\boldsymbol{\Theta}) = -\lambda_{\mu} \lVert \Laplace{\boldsymbol{\mu}} \rVert^2/2$. This added term on the log-likelihood acts as an attractive quadratic interactions of centres connected on the graph $\bm{A}$. This prior only impacts the M-step update of centre positions of Eq. \eqref{Mstep} as
\begin{equation} \label{eq:RGMM_mean_update}
    \boldsymbol{\mu}_k^{(t+1)} = \frac{\sum_{i=1}^N \boldsymbol{x}_i p_{ik}/\sigma^2 + 2\lambda_{\mu} \sum_{j=1}^K \boldsymbol{A}_{kj} \boldsymbol{\mu}^{(t+1)}_j}{\sum_{i=1}^N p_{ik}/\sigma^2 + 2\lambda_{\mu} \sum_{j=1}^K \boldsymbol{A}_{kj}}.
\end{equation}

As in Sect. \ref{sect:Soft}, it is possible to compute the value of $\sigma^2$ for which the high-temperature system becomes unstable, noted $T_c^\text{graph}$ considering perturbations around the fixed point $\boldsymbol{\mu}_k = \boldsymbol{0}_D$. Analogous derivations as in Sect. \ref{sect:Soft} shows that the system is unstable when the maximum eigenvalue of $\boldsymbol{M}$ is greater than 1, with
\begin{multline}
    \boldsymbol{M} = \left[ \left(\boldsymbol{I}_K - \frac{1}{K} \bm{J}_{K}\right) \otimes \bm{C} \right] \\
    \left[ \sigma^2 \boldsymbol{I}_{KD} + \frac{2\lambda_{\mu} K \sigma^4}{N} \boldsymbol{L} \otimes \boldsymbol{I}_D \right]^{-1},
\end{multline}
where $\boldsymbol{L}$ the Laplacian matrix defined as $\boldsymbol{L} = \boldsymbol{D} - \boldsymbol{A}$ with $\boldsymbol{D}$ the diagonal $K \times K$ degree matrix $\boldsymbol{D}_{kk} = \sum_i \boldsymbol{A}_{ik}$.
Since the prior is basically pulling adjacent centres, the threshold is translated toward lower values as $\lambda_{\mu}$ increases.

\begin{figure}
    \centering
    \subfigure{\includegraphics[width=1\linewidth]{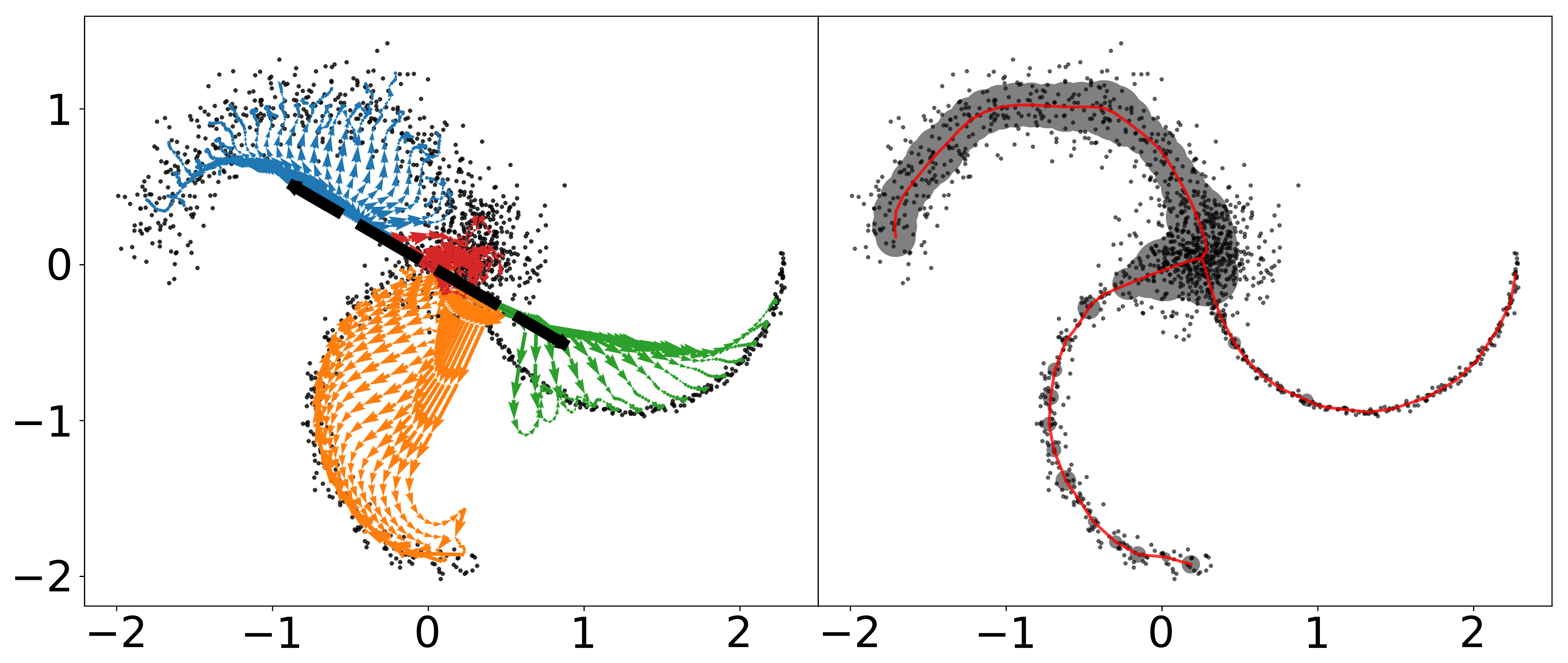}}
    
    
    \subfigure{\includegraphics[width=1\linewidth]{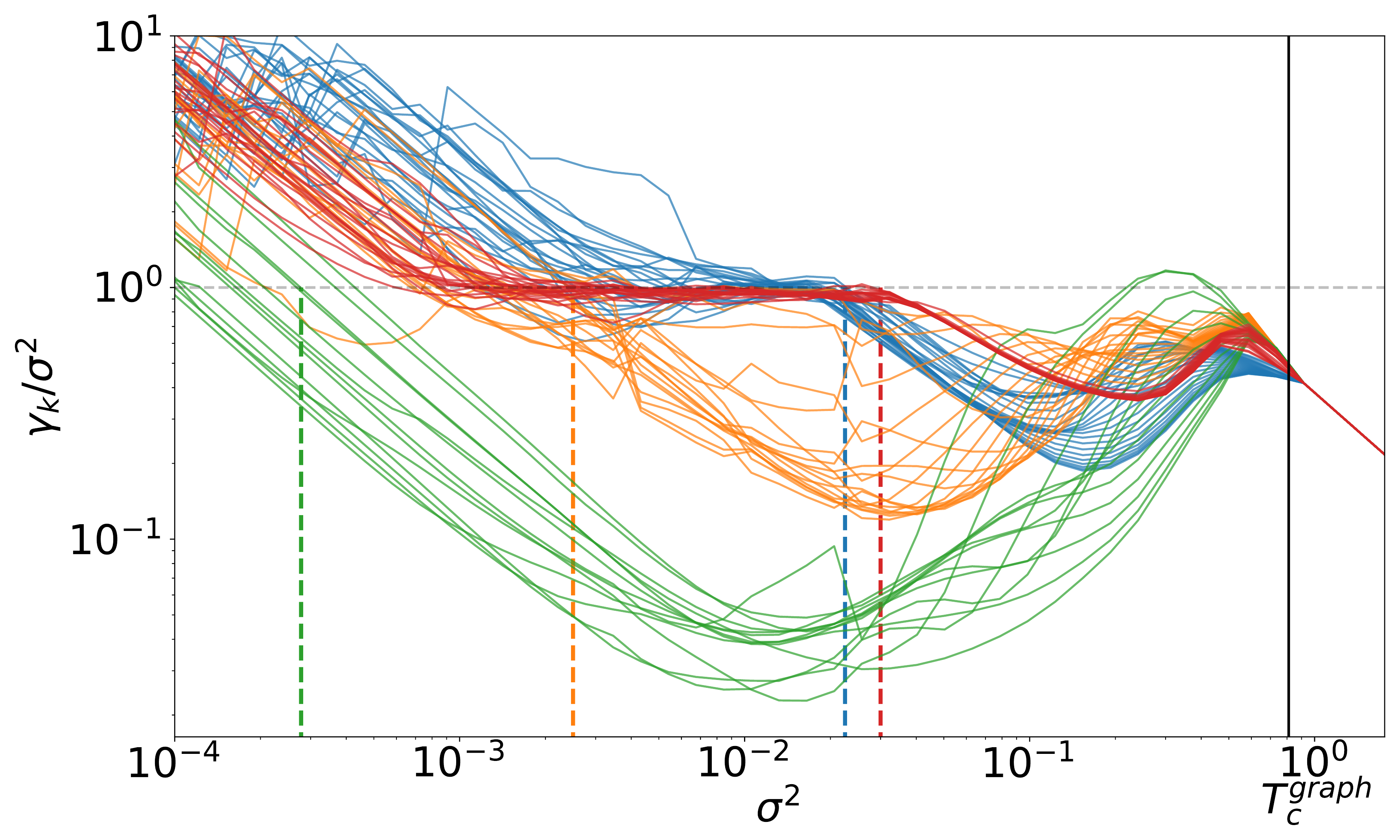}}
    
    \caption{\textit{(top left)} Displacement of $K=100$ components during the hard annealing of a tree branches dataset with different sampling standard deviations. Black dashed line corresponds to the first principal direction. Colours refer to branches in which centres end.
    \textit{(top right)} Learnt structure when stopping the annealing for components reaching the temperature $\sigma^2 \simeq \gamma_k$. Red lines are edges of the graph and grey shaded areas are 1-$\sigma_k$ circles.
    \textit{(bottom)} Evolution of the ratio $\gamma_k/\sigma^2$ as a function of $\sigma^2$. Vertical lines indicate the used variance for the generation of branches. The black vertical line corresponds to the value of $T_c^\text{graph}$.}
    \label{fig:Triforce_PT}
\end{figure}

When paving the data distribution with Gaussian clusters standing on a prior graph structure, the scale of interest is the local width of the elongated structure. This size is given locally, in our 2D case, by the minimum eigenvalue $\gamma_k$ of the weighted covariance $\boldsymbol{\Sigma}_k$. Figure \ref{fig:Triforce_PT} shows the result of a hard annealing procedure for $K=100$ components, $\lambda_{\mu}=300$ and using a graph prior given by the minimum spanning tree construction \citep{Boruvka1926}, hence assuming that centres are linked together with the minimum total length. This topological prior can takes different forms depending on the data under consideration and modifies the adjacency matrix $\boldsymbol{A}$ and, therefore, the value of $T_c^\text{graph}$.
As predicted by linear stability, centres are first aligned with the principal axis of the dataset at the beginning of the annealing and then spread over the structure to pave it more precisely, as shown on the top left panel. By tracking the evolution of the ratio $\gamma_k/\sigma^2$ in the bottom panel, we clearly distinguish four types of behaviours, signature of four distinct scales for structures in the dataset. It is also interesting to observe the absence of sharp phase transitions or splits within this continuous dataset tending to smooth out the evolution of the energy when the temperature is decreasing.
By imposing, for each component, the variance $\sigma^2$ during the annealing at the moment $\gamma_k \simeq \sigma^2$, we obtain the graph of the top right panel, showing multiple adaptive scales, even though branches have one order of magnitude difference in sampling standard deviation.

\section{Conclusions}

In this work, we exhibited the cascade of phase transitions occurring when tracking the evolution of eigenvalues of the successive covariance matrices of mixture components during a simulated annealing of the EM algorithm in multiple cases. We saw that hard annealing has nice theoretical advantages with the ability to compute successive theoretical transitions and we showed that a relaxed version can provide an accurate description for datasets exhibiting multiple scales with nested clusters. We also illustrated how hard annealing can be used jointly with graph regularisation of GMs to learn a multi-scale principal graph independently on the initialisation.
By revisiting the widely known problem of unsupervised clustering formulated in the framework of statistical physics, the proposed way of exploring the data through simulated annealing does not lead to an automatic and blind solution using all the inputted components. Instead, it builds a hierarchical description independent from $K$ providing a qualitative and quantitative insight on the structure of the dataset at different scales and without requiring any prior knowledge.
The 2D diagram allows the analysis of the annealing independently from the data dimensionality and highlights characteristic scales at which physical transitions occur.
These information on the number of components, their scale and hierarchy can be used a posteriori for data exploration before running blind clustering methods.

Interestingly, we saw that the latent variables of the GMM are directly related to the values taken by the order parameter. Since the GMM can be recast into a Restricted Boltzmann Machine \citep{Hinton2002} using a soft-max prior on the hidden nodes, making it a particular type of autoencoder \citep{Bourlard1998}, our work can be seen as the learning of a latent representation of the phase transitions, as is discussed in \citep{VanNieuwenburg2017, Wetzel2017}.

\bigskip

\paragraph*{Acknowledgments.} This research has been supported by the funding for the ByoPiC project from the European Research Council (ERC) under the European Union’s Horizon 2020 research and innovation program grant number ERC-2015-AdG 695561. A.D. was supported by the Comunidad de Madrid and the Complutense University of Madrid (Spain) through the Atracción de Talento program (Ref. 2019-T1/TIC-13298).

\bibliography{GMM_stat}

\end{document}